\documentclass[10pt,twocolumn,letterpaper]{article}

\usepackage[pagenumbers]{cvpr} 

\usepackage{mathtools}
\usepackage{algorithm}
\usepackage{algpseudocode}
\usepackage{overpic}

\usepackage{tkz-euclide}
\usepackage{tikz}
\usepackage{pgfplots}
\pgfplotsset{compat=1.14}
\usetikzlibrary{positioning,calc,fadings,backgrounds,fit,3d,shapes.misc,shapes.geometric}
\usetikzlibrary{matrix}

\usepackage[T1]{fontenc}

\usepackage{amsmath}
\usepackage{amssymb}

\DeclareMathOperator{\E}{\mathbb{E}}

\definecolor{target}{RGB}{128,0,128}
\definecolor{source}{RGB}{255, 169, 0}

\definecolor{amethyst}{rgb}{0.6, 0.4, 0.8}

\definecolor{applegreen}{rgb}{0.55, 0.71, 0.0}
\definecolor{apricot}{rgb}{0.98, 0.81, 0.69}
\definecolor{aqua}{rgb}{0.0, 1.0, 1.0}
\definecolor{atomictangerine}{rgb}{1.0, 0.6, 0.4}
\definecolor{azure}{rgb}{0.0, 0.5, 1.0}
\definecolor{beaublue}{rgb}{0.74, 0.83, 0.9}
\definecolor{beaver}{rgb}{0.62, 0.51, 0.44}

\newcommand*{\dittostraight}{---\textquotedbl---}

\definecolor{cvprblue}{rgb}{0.21,0.49,0.74}
\usepackage[pagebackref,breaklinks,colorlinks,citecolor=cvprblue]{hyperref}

\title{Functional Diffusion}

\author{Biao Zhang\\
KAUST\\
{\tt\small biao.zhang@kaust.edu.sa}
\and
Peter Wonka\\
KAUST\\
{\tt\small pwonka@gmail.com}
}

\begin{document}
\twocolumn[{
	\renewcommand\twocolumn[1][]{#1}%
	\maketitle
	\begin{center}
		\captionsetup{type=figure}
        \setlength{\abovecaptionskip}{-0.001cm}
		\includegraphics[width=\textwidth]{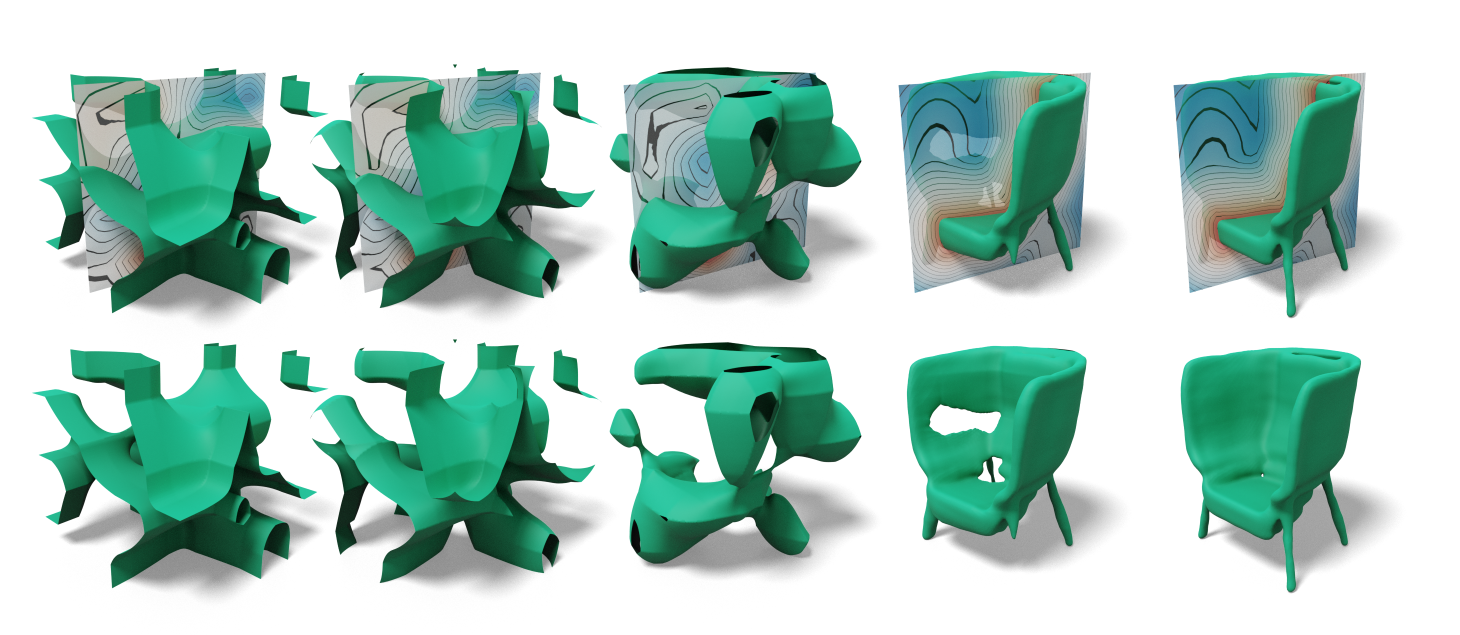}
            \vspace{-15pt}
		\captionof{figure}{\textbf{Functional diffusion.} Our method is able to generate complicated functions with a continuous domain. From left to right, we show 5 steps of the generating process. This particular example shows signed distance functions and we show the zero-isosurface of the generated function in green. Furthermore, we visualize the function values on a plane, where the red colors mean larger and blue means smaller.}
		\label{fig:teaser}
	\end{center}
}]
\begin{abstract}
We propose a new class of generative diffusion models, called functional diffusion. In contrast to previous work, functional diffusion works on samples that are represented by functions with a continuous domain. Functional diffusion can be seen as an extension of classical diffusion models to an infinite-dimensional domain. Functional diffusion is very versatile as images, videos, audio, 3D shapes, deformations, \etc, can be handled by the same framework with minimal changes. In addition, functional diffusion is especially suited for irregular data or data defined in non-standard domains. In our work, we derive the necessary foundations for functional diffusion and propose a first implementation based on the transformer architecture. We show generative results on complicated signed distance functions and deformation functions defined on 3D surfaces.
\end{abstract}

\section{Introduction}
In the last two years diffusion models have become the most popular method for generative modeling of visual data, such as 2D images~\cite{rombach2022high,saharia2022photorealistic}, videos~\cite{ho2022imagen, ho2022video}, and 3D shapes~\cite{cheng2023sdfusion, zheng2023locally, 10.1145/3592442, hui2022neural}.
In order to train a diffusion model, one needs to add and subtract noise from a data sample.
In order to represent a sample, many methods use a direct representation, such as a 2D or 3D grid. Since diffusion can be very costly, this representation is often used in conjunction with a cascade of diffusion models~\cite{ho2022cascaded, saharia2022photorealistic}.
Alternatively, diffusion methods can represent samples in a compressed latent space~\cite{rombach2022high}. A sample can be encoded and decoded to the compressed space using an autoencoder whose weights are trained in a separate pre-process.

\begin{figure}
    \centering
    \begin{overpic}[trim={1cm 0cm 1cm 3cm},clip,
    width=\linewidth,
    grid=false]{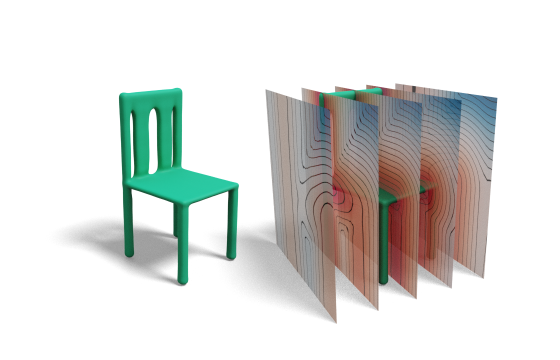}
    \end{overpic}
    \vspace{-20pt}
    \caption{\textbf{Signed distance functions.} We show a 3D shape on the left and on the right, we visualize the signed distances sampled in several parallel planes.}
    \label{fig:sdf}
\end{figure}

In our work, we explore a departure from these previous approaches and set out to study diffusion in a functional space. We name the resulting method \emph{functional diffusion}. 
In functional diffusion, the data samples are functions in a function space (see an example function in~\cref{fig:sdf}). 
In contrast to regular diffusion, we do not start with a noisy sample, but we need to define a noise function as a starting point. This noise function is then gradually denoised to obtain a sample from the function space. In order to realize our idea we need multiple different representations that are different from regular diffusion.
We employ both a continuous and a sampled representation of a function.
As the continuous representation of a function, we propose a set of vectors that are latent vectors of a functional denoising network.
To represent a sampled function, we use a set of point samples in the domain of the function together with the corresponding function values.
Both of these function representations are used during training and inference. 
The method is initialized by sampling a noisy continuous function that spans the complete domain.
Then we evaluate this function at discrete locations to obtain a sampled representation.
A training step in functional diffusion takes both the continuous and sampled representation as input and tries to predict a new continuous representation that is a denoised version of the input function.
This novel form of diffusion has multiple interesting properties. First, the framework is very versatile and can be directly adapted to many different forms of input data. We can handle images, videos, audio, 3D shapes, deformations, \etc, with the same framework.
Second, we can directly handle irregular data and non-standard domains as there are few constraints on the function domain as well as the samples of the sampled function representation. For example, we can work with deformations on a surface, which constitutes an irregular domain.
Third, we can decouple the representational power of the continuous and sampled function representation.
Finally, we believe the idea of functional diffusion is inherently technically interesting. It is a non-trivial change and our work can lay the foundation for a new class of diffusion models with many variations.

In summary, we make the following major contributions:
\begin{itemize}
\item We introduce the concept of functional diffusion, explain the technical background and derive the corresponding equations.
\item We propose a technical realization and implementation of the functional diffusion concept.
\item We demonstrate functional diffusion on irregular domains that are challenging to handle for existing diffusion methods.
\item We demonstrate improved results on shape completion from sparse point clouds.
\end{itemize}

\section{Related Work}
\subsection{Generative Models}
Generative models have been extensively explored for image data. We have seen several popular generative models in past years such as Generative Adversarial Networks (GANs)~\cite{goodfellow2014generative}, Variational Autoencoders (VAEs)~\cite{DBLP:journals/corr/KingmaW13} and Diffusion Probabilistic Models (DPMs)~\cite{ho2020denoising}. GANs utilize an adversarial training process. The versatility in generating high-dimensional data has been proven by numerous applications and improvements. VAEs aim to learn a representation space of the data with an autoencoder and enable the generation of new samples by sampling from the learned space. However, the quality is often lower than GANs. This idea is further improved in DPMs. Instead of decoding the representation with a one-step decoder, DPMs developed a new mechanism of progressive decoding. 
DPMs have demonstrated remarkable success in capturing and generating complex patterns in image data~\cite{ho2020denoising, ho2022cascaded, saharia2022photorealistic, rombach2022high}.

\subsection{Diffusion probabilistic models}
When DPMs were invented in the beginning, they showed significant advantages in generating quality and diversity. However, the disadvantages are also obvious. For example, the sampling process is slower than other generative models. Some works~\cite{song2020denoising, lu2022dpm, karras2022elucidating, lu2022dpmpp} are dedicated to solving the slow sampling problem. On the other hand, these works~\cite{bansal2022cold, rissanen2022generative} are proposed to solve the cases of non-Gaussian noise/degradation. However, our focus is to propose a new diffusion model for functional data. Common data forms like images can be seen as lying in a finite-dimensional space. However, a function is generally infinite-dimensional. It is not straightforward to adapt existing diffusion models for functional data. A direct solution is a two-stage training method. The first stage is
to fit a network to encode functions with finite-dimensional latent space. In the later stage, a generative diffusion model is trained in the learned latent space. Many methods follow this design~\cite{cheng2023sdfusion,10.1145/3592442, muller2023diffrf}. On the other hand, SSDNerf~\cite{ssdnerf} combines both stages into one that jointly optimizes an autodecoder and a latent diffusion model. However, the method still trains diffusion in the latent space.
The most related work to our proposed method is DPF~\cite{zhuang2023diffusion}. However, DPF still works on data sampled on a discrete grid. Thus the generated sample is still defined in a fixed resolution. We refer the reader to a recent survey of diffusion models in various domains besides images~\cite{Po:2023:star_diffusion_models}.


\subsection{Neural Fields}
Neural networks are often used to represent functions with a continuous domain. Here are some types of neural field applications: 1) in computer graphics and geometry processing, 3D shapes can be represented with implicit functions and thus are suitable to be modeled with neural networks~\cite{mescheder2019occupancy, park2019deepsdf, peng2020convolutional, li2022learning, yan2022shapeformer, tang2021sa, cheng2023sdfusion, 10.1145/3592442}; 2) 3D textured objects and scenes can be rendered with radiance fields~\cite{mildenhall2021nerf} which are also modeled with MLPs; 3) in physics, researchers use neural networks to represent complex functions which serve as solutions of differential equations~\cite{raissi2017physics}. Because of the universal approximation ability of neural networks, neural fields often provide flexibility in handling complex and high-dimensional data, and they can be trained end-to-end using gradient-based optimization techniques. Most importantly, neural fields can hold data sampled from an infinite large resolution. We refer the reader to a recent survey for more details on neural fields~\cite{10.1111:cgf.14505}.

\section{Methodology}
\begin{figure}[tb]
    \centering
    \centering
    \begin{overpic}[trim={4cm 4cm 4cm 0cm},clip,
    width=\linewidth,
    grid=false]{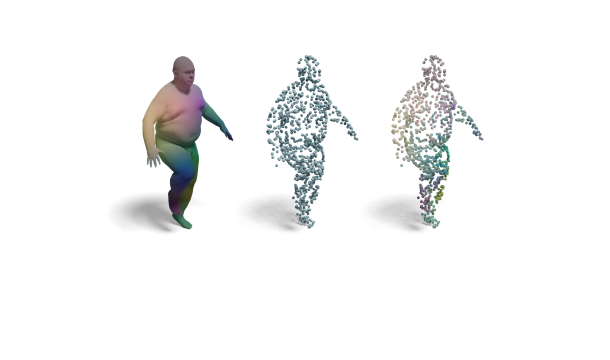}
        \put(15, 53){\small{$f$}}
        \put(45, 53){\small{$\{\mathbf{x}_i\}_{i\in\mathcal{C}}$}}
        \put(73, 53){\small{$\{\mathbf{x}_i, f(\mathbf{x}_i)\}_{i\in\mathcal{C}}$}}
    \end{overpic}
    \vspace{-10pt}
    \caption{\textbf{Function approximation.} We illustrate how to approximate a function with its discretized state. 
    Left: a function whose domain $\mathcal{X}$ is a manifold. Middle: sampled points in the domain. Right: the sampled points and the corresponding function values. Different from DPMs which sample on a grid of a fixed resolution, we do not have this restriction.}
    \label{fig:func-approx}
\end{figure}

We first introduce the definition of the functional diffusion in~\cref{sec:problm}. Then we show how to train the denoising network in~\cref{sec:param}. Lastly, we show how we sample a function from the trained functional diffusion models in~\cref{sec:infer}.
\begin{table}[]
    \centering
    \setlength{\tabcolsep}{1pt}
    \begin{tabular}{cc}
    \toprule
    DPM & Proposed \\ \midrule
    Gaussian $\mathbf{n}\in\mathbb{R}^{C}$ & Random function $g:\mathcal{X} \rightarrow \mathcal{Y}$ \\ 
    Finite-dim $\mathbf{x}_0\in\mathbb{R}^{C}$& Infinite-dim $f_0:\mathcal{X} \rightarrow \mathcal{Y}$ \\
    Noised $\mathbf{x}_t\in\mathbb{R}^{C}$ & Noised function $f_t:\mathcal{X} \rightarrow \mathcal{Y}$ \\
    \dittostraight & Context $\{\mathbf{x}_i, f_t(\mathbf{x}_i)\}_{i\in\mathcal{Q}}$\\
    \dittostraight & Queries $\{\mathbf{x}_i, f_t(\mathbf{x}_i)\}_{i\in\mathcal{C}}$\\
    $D_\theta:\mathbb{R}^C\rightarrow \mathbb{R}^C$ & $D_\theta:\{f:\mathcal{X} \rightarrow \mathcal{Y}\}\rightarrow \{f:\mathcal{X} \rightarrow \mathcal{Y}\}$
         \\ \bottomrule
    \end{tabular}
    \caption{\textbf{Comparison of classical DPMs and the proposed method.} For DPMs, the data samples are finite-dimensional and the denoiser is a function of the noised data $\mathbf{x}_t$. Our method deals with infinite-dimensional functions with a continuous domain. Thus the denoiser $D_\theta$ is becoming a ``function of a function''. This inspires us to seek a solution to find a way to process infinite-dimensional functions with neural networks. Also note that DPM is a special case when $\mathcal{Q}=\mathcal{C}$.}
    \label{tab:comp-dpm-proposed}
\end{table}
\subsection{Problem Definition}\label{sec:problm}
The training dataset $\mathcal{D}$ contains a collection of functions $\textcolor{target}{f_0}$ with continuous domains $\mathcal{X}$ and range $\mathcal{Y}$,
\begin{equation}
    \textcolor{target}{f_0}:\mathcal{X} \rightarrow \mathcal{Y}.
\end{equation}
For example, we can represent watertight meshes as signed distance functions $\textcolor{target}{f_0}:\mathbb{R}^{3}\rightarrow \mathbb{R}^{1}$.
We also define function set $\mathcal{F}$ where each element is also a function
\begin{equation}
    \textcolor{source}{g}:\mathcal{X} \rightarrow \mathcal{Y}.
\end{equation}
The function $\textcolor{source}{g}$ works similarly to the noise in traditional diffusion models. However, in functional diffusion, we require the ``noise'' to be a function.
We can obtain a ``noised'' version $f_t$ given $f_0$ from $\mathcal{D}$ and $g$ from $\mathcal{F}$,
\begin{equation}\label{eq:noised}
    f_t(\mathbf{x}) = \alpha_t \cdot \textcolor{target}{f_0}(\mathbf{x})+ \sigma_t \cdot \textcolor{source}{g}(\mathbf{x}),
\end{equation}
where $t$ is a scalar from $0$ (\textcolor{target}{least noisy}) to $1$ (\textcolor{source}{most noisy}). We name $f_t$ as the \emph{noised state} at timestep $t$. The terms $\alpha_t$ and $\sigma_t$ are positive scalars. In DDPM~\cite{ho2020denoising}, they satisfy $\alpha_t^2 + \sigma_t^2=1$. Thus $\alpha_t$ is a monotonically decreasing function of $t$, while $\sigma_t$ is monotonically increasing. VDM~\cite{kingma2021variational} characterizes $\alpha_t^2/\sigma_t^2$ as signal-to-noise ratio (SNR).

Our goal is to train a denoiser which can approximate:
\begin{equation}\label{eq:denoiser-approximate}
    D_\theta[f_t, t](\mathbf{x})\approx \textcolor{target}{f_0}(\mathbf{x}).
\end{equation}
This is often called $x_0$-prediction~\cite{kingma2021variational} in the literature of diffusion models. However, other loss objectives also exist, \eg, $\epsilon$-prediction~\cite{ho2020denoising}, $v$-prediction~\cite{salimans2022progressive} and $f$-prediction~\cite{karras2022elucidating}. We emphasize that choosing $x_0$-prediction is important in the proposed functional diffusion which will be explained later.

The objective is
\begin{equation}\label{eq:diffusion-obj}
    \E_{\textcolor{target}{f_0}\in\mathcal{D}, \textcolor{source}{g}\in\mathcal{F}, t\sim T(t)}\left[w(t)d \left(D_\theta[f_t, t],  \textcolor{target}{f_0}\right)^2\right],
\end{equation}
where $d(\cdot, \cdot)$ is a metric defined on the function space $\{f:\mathcal{X}\rightarrow\mathcal{Y}\}$ and $w(t)$ is a weighting term. We summarize the differences between the vanilla DPMs and the proposed functional diffusion in~\cref{tab:comp-dpm-proposed}.

\begin{algorithm}[tb]
\caption{Training}\label{alg:train}
\begin{algorithmic}[1]
\Repeat 
    \State $\textcolor{source}{g}\in\mathcal{F}$ \Comment{noise function}
    \State $\textcolor{target}{f_0}\in\mathcal{D}$ \Comment{training function}
    \State $t\sim\mathcal{T}$ \Comment{noise level}
    \State $\alpha_t=1/\sqrt{t^2+1}$, $\sigma_t=t/\sqrt{t^2+1}$ \Comment{SNR}
    \State Sample $\mathcal{C}$ \Comment{context}
    \State Evaluate $\{g(\mathbf{x}_i)\}_{i\in\mathcal{C}}$ and $\{f_0(\mathbf{x}_i)\}_{i\in\mathcal{C}}$
    \State Calculate the context $\{f_t(\mathbf{x}_i)\}_{i\in\mathcal{C}}$ with Eq.~\eqref{eq:noised}
    \State Sample $\mathcal{Q}$ \Comment{query}
    \State Optimize Eq.~\eqref{eq:network-denoise} \Comment{denoise}
\Until convergence
\end{algorithmic}
\end{algorithm}

\begin{algorithm}[tb]
\caption{Sampling}\label{alg:sample}
\begin{algorithmic}[1]
\Ensure Sample $\mathcal{C}$ and $\textcolor{source}{g}\in\mathcal{F}$ 
\State Let $f_t = \textcolor{source}{g}$
\State Evaluate $\{\mathbf{x}_i, f_t(\mathbf{x}_i)\}_{i\in\mathcal{C}}$
\For {$k \in \{N, N-1, \dots, 2, 1\}$}
    \State $t_{k} = T(k)$, $t_{k-1} = T(k-1)$
    \State $\alpha_t=1/\sqrt{t_k^2+1}$, $\alpha_s=1/\sqrt{t_{k-1}^2+1}$
    \State $\sigma_t=t_k/\sqrt{t_k^2+1}$, $\sigma_s=t_{k-1}/\sqrt{t_{k-1}^2+1}$
    \State Predict $\{f_s(\mathbf{x}_i)\}_{i\in\mathcal{C}}$ with Eq.~\eqref{eq:sampling}
    \State Let $f_t \leftarrow f_s$
\EndFor
\State $f_0(\mathbf{x})=D_\theta\left(\{\mathbf{x}_i, f_t(\mathbf{x}_i)\}_{i\in\mathcal{C}}, t, \mathbf{x}\right)$
\end{algorithmic}
\end{algorithm}

\begin{table*}[]
\centering
\setlength{\tabcolsep}{3pt}
\begin{tabular}{cccccccc}
\toprule
 Task & Input & Domain Space & Output & Range Space & Condition & $|\mathcal{C}|$ & $|\mathcal{Q}|$\\ \midrule
3D Shapes & Coordinates & $\mathbb{R}^3$ & SDF & $\mathbb{R}^1$ & Surface Point Clouds & 49152 & 2048\\
3D Deformation & Points on Manifold & $\mathcal{M}$ & Vector & $\mathbb{R}^3$ & Sparse Correspondence & 16384 & 2048
\\
\bottomrule
\end{tabular}
\caption{\textbf{Task designs.} We show the two main tasks used to prove the efficiency of the proposed method.}
\label{tab:task-parameters}
\end{table*}

\subsection{Parameterization}\label{sec:param}

\begin{figure}[tb]
    \centering
\begin{tikzpicture}[>=Stealth, tight background, remember picture, font=\normalfont]

\tikzset{auto matrix/.style={matrix of nodes,
inner sep=0pt, ampersand replacement=\&,
nodes in empty cells,column sep=1.0pt,row sep=0.4pt,
}}

\tikzstyle{layer} = []

\matrix[auto matrix=m,xshift=0em,yshift=0em,opacity=0.9, 
    row 1/.style={nodes={draw=beaver!80, fill=beaver!20}},
    cells={nodes={minimum width=0.8em,minimum height=0.8em,
    very thin,anchor=center,
    }},
    label=above:{$f_1(\mathbf{x})$},
    ](f1){
    \\
};
\matrix[right=1.7cm of f1, auto matrix=m,xshift=0em,yshift=0em,opacity=0.9, 
    row 1/.style={nodes={draw=beaver!80, fill=beaver!30}},
    cells={nodes={minimum width=0.8em,minimum height=0.8em,
    very thin,anchor=center,
    }},
    label=above:{$f_t(\mathbf{x})$},
    ](ft){
    \\
};
\matrix[right=1.7cm of ft, auto matrix=m,xshift=0em,yshift=0em,opacity=0.9, 
    row 1/.style={nodes={draw=beaver!80, fill=beaver!40}},
    cells={nodes={minimum width=0.8em,minimum height=0.8em,
    very thin,anchor=center,
    }},
    label=above:{$f_s(\mathbf{x})$},
    ](fs){
    \\
};
\matrix[right=1.7cm of fs, auto matrix=m,xshift=0em,yshift=0em,opacity=0.9, 
    row 1/.style={nodes={draw=beaver!80, fill=beaver!50}},
    cells={nodes={minimum width=0.8em,minimum height=0.8em,
    very thin,anchor=center,
    }},
    label=above:{$f_0(\mathbf{x})$},
    ](f0){
    \\
};

\draw[->] (f1) -- (ft);
\draw[->] (ft) -- (fs);
\draw[->, dashed] (fs) -- (f0);

\matrix[below=0.2cm of f1,auto matrix=m,xshift=0em,yshift=0em,opacity=0.9, 
    row 1/.style={nodes={draw=applegreen!80, fill=applegreen!20}},
    row 2/.style={nodes={draw=apricot!80, fill=apricot!20}},
    row 3/.style={nodes={draw=aqua!80, fill=aqua!20}},
    row 4/.style={nodes={draw=atomictangerine!80, fill=atomictangerine!20}},
    row 5/.style={nodes={draw=azure!80, fill=azure!20}},
    cells={nodes={minimum width=0.8em,minimum height=0.8em,
    very thin,anchor=center,
    }},
    label=below:{$\{f_1(\mathbf{x}_i)\}_{i\in\mathcal{C}}$},
    ](f1_c){
    \\ \\ \\ \\
};
    
\matrix[below=0.2cm of ft,auto matrix=m,xshift=0em,yshift=0em,opacity=0.9, 
    row 1/.style={nodes={draw=applegreen!80, fill=applegreen!30}},
    row 2/.style={nodes={draw=apricot!80, fill=apricot!30}},
    row 3/.style={nodes={draw=aqua!80, fill=aqua!30}},
    row 4/.style={nodes={draw=atomictangerine!80, fill=atomictangerine!30}},
    row 5/.style={nodes={draw=azure!80, fill=azure!30}},
    cells={nodes={minimum width=0.8em,minimum height=0.8em,
    very thin,anchor=center,
    }},
    label=below:{$\{f_t(\mathbf{x}_i)\}_{i\in\mathcal{C}}$},
    ](ft_c){
    \\ \\ \\ \\
};

\matrix[below=0.2cm of fs,auto matrix=m,xshift=0em,yshift=0em,opacity=0.9, 
    row 1/.style={nodes={draw=applegreen!80, fill=applegreen!40}},
    row 2/.style={nodes={draw=apricot!80, fill=apricot!40}},
    row 3/.style={nodes={draw=aqua!80, fill=aqua!40}},
    row 4/.style={nodes={draw=atomictangerine!80, fill=atomictangerine!40}},
    row 5/.style={nodes={draw=azure!80, fill=azure!40}},
    cells={nodes={minimum width=0.8em,minimum height=0.8em,
    very thin,anchor=center,
    }},
    label=below:{$\{f_s(\mathbf{x}_i)\}_{i\in\mathcal{C}}$},
    ](fs_c){
    \\ \\ \\ \\
};
\matrix[below=0.2cm of f0,auto matrix=m,xshift=0em,yshift=0em,opacity=0.9, 
    row 1/.style={nodes={draw=applegreen!80, fill=applegreen!50}},
    row 2/.style={nodes={draw=apricot!80, fill=apricot!50}},
    row 3/.style={nodes={draw=aqua!80, fill=aqua!50}},
    row 4/.style={nodes={draw=atomictangerine!80, fill=atomictangerine!50}},
    row 5/.style={nodes={draw=azure!80, fill=azure!50}},
    cells={nodes={minimum width=0.8em,minimum height=0.8em,
    very thin,anchor=center,
    }},
    label=below:{$\{f_0(\mathbf{x}_i)\}_{i\in\mathcal{C}}$},
    ](f0_c){
    \\ \\ \\ \\
};

\draw[->] (f1_c) -- (ft_c);
\draw[->] (ft_c) -- (fs_c);
\draw[->] (fs_c) -- (f0_c);

\draw[->] (f1_c) -- (ft);
\draw[->] (ft_c) -- (fs);
\draw[->] (fs_c) -- (f0);

\end{tikzpicture}
    \caption{\textbf{Inference chain.} We show a simplified $4$-steps generating process in~\cref{eq:sampling}. The arrows show how the data flows during inference. $\mathbf{x}$ represents an arbitrary query coordinate. $\mathcal{C}$ is the context set. The state $f_s(\mathbf{x})$ requires to know both the previous state $f_t(\mathbf{x})$ and $\{f_t(\mathbf{x}_i)\}_{i\in\mathcal{C}}$. Thus it is dependent on all previous states $f_{<s}$. $f_0(\mathbf{x})$ is the only exception because $\sigma_0=0$. Thus $f_0(\mathbf{x})$ is fully decided by the penultimate state $\{f_s(\mathbf{x}_i)\}_{i\in\mathcal{C}}$.}
    \label{fig:inference-chain}
\end{figure}
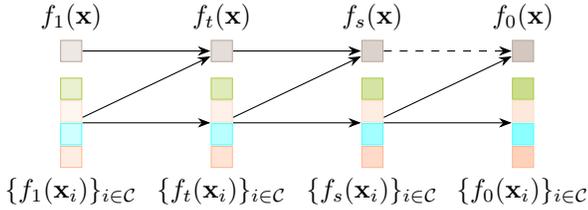

\begin{figure*}[!htbp]
    \centering
    \input{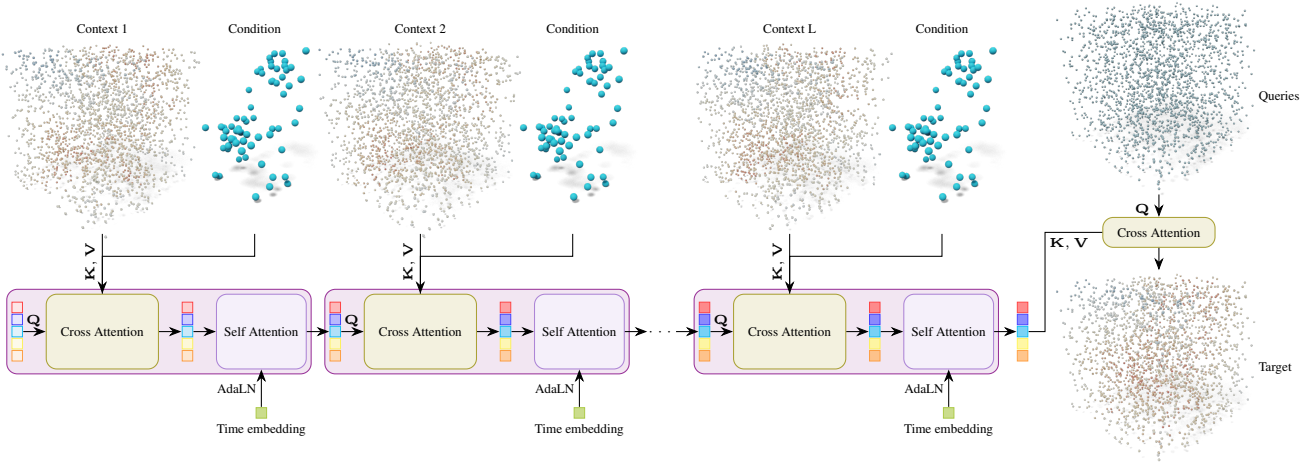}
    \vspace{-10pt}
    \caption{\textbf{The network design of the SDF diffusion model.} The context set is split into $L$ smaller ones. They (and optionally conditions such as sparse surface point clouds) are fed into different stages of the network by using cross-attention. The time embedding is injected into the network in every self-attention layer by adaptive layer normalization. After $L$ stages, we obtain the representation vector sets and they will be used to predict values of arbitrary queries. For SDFs, we optimize simple minimum squared errors.
    }
    \label{fig:sdf-pipeline}
\end{figure*}

\paragraph{Denoising network.} The functional $D_\theta$ is parameterized by a neural network $\theta$. It is impossible to feed the noised state function $f_t$ directly to the neural network as input. In order to make the computation tractable, our idea is to represent functions with a set of coordinates together with their corresponding values. Thus we sample (discretize) a set $\{\mathbf{x}_i \in \mathcal{X}\}_{i\in\mathcal{C}}$ in the domain $\mathcal{X}$ of $f_t$. We feed this set to the denoising network along with the corresponding function values $\{f_t(\mathbf{x}_i)\}_{i\in\mathcal{C}}$ (also see~\cref{fig:func-approx} for an illustration),
\begin{equation}\label{eq:approx-denoising-network}
D_\theta[f_t, t](\mathbf{x})\approx D_\theta\left(\{\mathbf{x}_i, f_t(\mathbf{x}_i)\}_{i\in\mathcal{C}}, t, \mathbf{x}\right).
\end{equation}
The design of the network $D_\theta$ varies for different applications. However, we give a template design in later sections.
\paragraph{Function metric.}
For the function metric $d(\cdot, \cdot)$, we choose the $l$-2 metric,
\begin{equation}
    d(D_\theta[f_t, t], \textcolor{target}{f_0}) = \left(\int_\mathcal{X} \left|D_\theta[f_t, t](\mathbf{x}) - \textcolor{target}{f_0}(\mathbf{x})\right|^2 \mathrm{d}\mathbf{x} \right)^{1/2}
\end{equation}
The approximation of the metric $d(\cdot, \cdot)$ is also done by sampling (Monte-Carlo integration), 
\begin{equation}
    d\left(D_\theta[f_t, t],  \textcolor{target}{f_0}\right) \approx \left(\sum_{i\in\mathcal{Q}}\left| D_\theta[f_t, t](\mathbf{x}_i) - \textcolor{target}{f_0}(\mathbf{x}_i) \right|^2\right)^{1/2}.
\end{equation}
Thus our loss objective in Eq.~\eqref{eq:diffusion-obj} can be written as, 
\begin{equation}\label{eq:network-denoise}
    w(t)\sum_{i\in\mathcal{Q}}\left| D_\theta\left(\{\mathbf{x}_j, f_t(\mathbf{x}_j)\}_{j\in\mathcal{C}}, t, \mathbf{x}_i\right) - \textcolor{target}{f_0}(\mathbf{x}_i) \right|^2.
\end{equation}
Pixel diffusion (DPMs trained in the pixel space) can be seen as a special case of the model by sampling $\mathcal{C}$ on a fixed regular grid and letting $\mathcal{Q}=\mathcal{C}$. DPF~\cite{zhuang2023diffusion} uses the term \emph{context} for $\mathcal{C}$ and \emph{query} for $\mathcal{Q}$. Thus we also follow this convention. We summarized how we design $\mathcal{Q}$ and $\mathcal{C}$ for different tasks in~\cref{tab:task-parameters}.
\paragraph{Initial noise function.}
For now, we still do not know how to choose the noise function set $\mathcal{F}=\{\textcolor{source}{g}:\mathcal{X}\rightarrow\mathcal{Y}\}$. In DPMs, the noise is often modeled with a standard Gaussian distribution. Gaussian processes are an infinite-dimensional generalization of multivariate Gaussian distributions. Thus, it is straightforward to use Gaussian processes to model the noise functions. However, in our practical experiments, we find sampling from Gaussian processes is time-consuming during training. Thus, we choose a simplified version. 
In the case of Euclidean space,
we sample Gaussian noise on a grid in $\mathcal{X}$. Then other values are interpolated with the values on the grid. If the domain $\mathcal{X}$ is a non-Euclidean manifold which is difficult to sample, instead we define the noise function in the ambient space of $\mathcal{X}$.
In this way, we defined a way to build the function set $\mathcal{F}$. During training, in each iteration, we sample a noise function $g$ from this set.

To sum up, the training algorithm can be found in~\cref{alg:train}.

\begin{figure}[tb]
    \centering
    \input{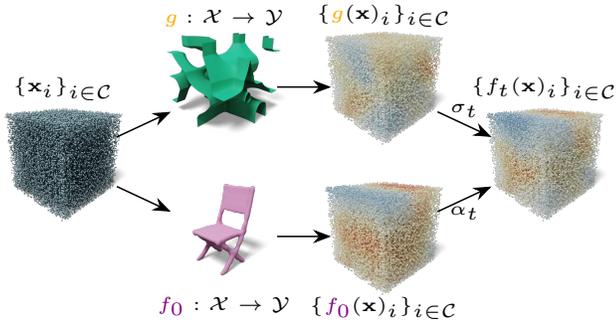}
    \vspace{-10pt}
    \caption{\textbf{Evaluation of the context $\{\mathbf{x}_i, f_t(\mathbf{x}_i)\}_{i\in\mathcal{C}}$.} We sample a set of points $\{\mathbf{x}_i\}_{i\in\mathcal{C}}$ in the domain $\mathcal{X}$. We evaluate the values both in the noise function $\textcolor{source}{g}$ and the ground-truth function $\textcolor{target}{f_0}$. This is how~\cref{eq:noised} works.
    }
    \label{fig:noised_states}
\end{figure}

\subsection{Inference}\label{sec:infer}


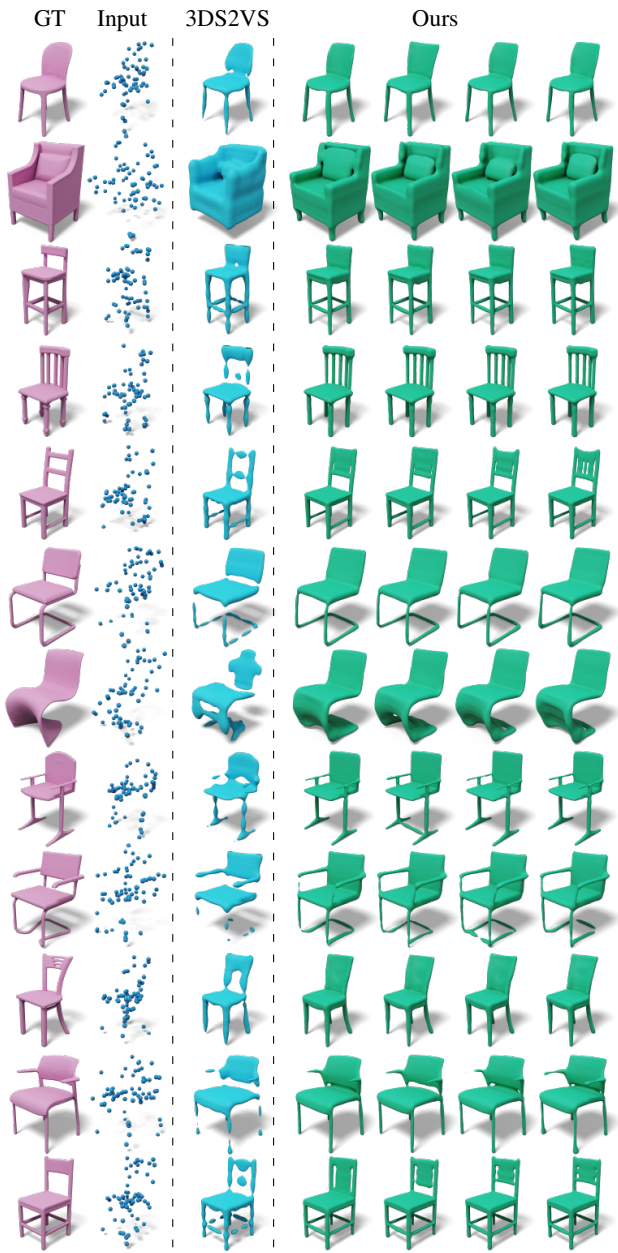
\begin{figure}[!htbp]
    \centering
    \begin{overpic}[trim={1cm 0cm 1cm 0cm},clip,
    width=\linewidth,
    grid=false]{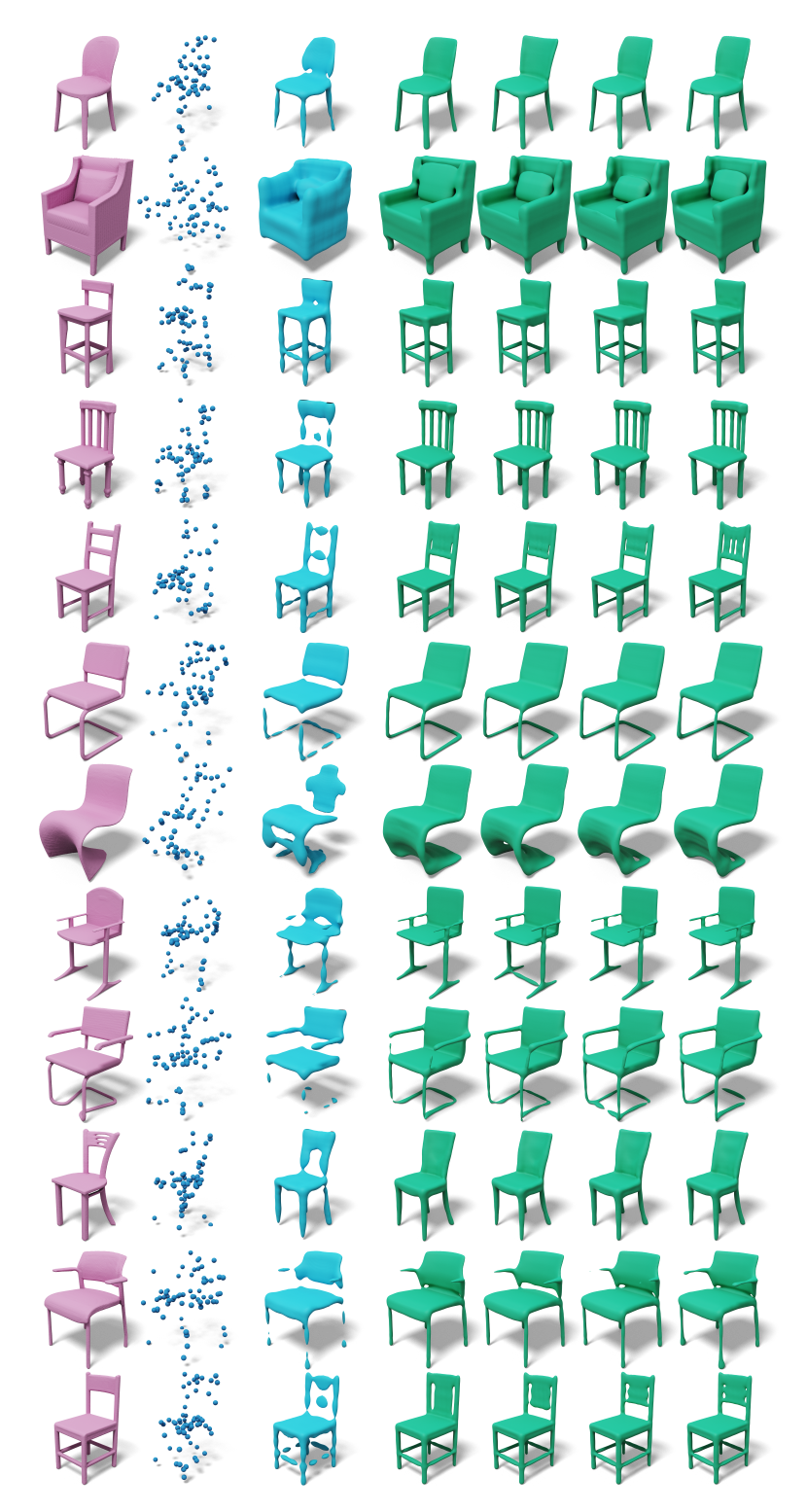}
        \dashline{0.7}(14,2)(14,98)
        \dashline{0.7}(22,2)(22,98)
        \put(3, 99){\small{GT}}
        \put(8, 99){\small{Input}}
        \put(15, 99){\small{3DS2VS}}
        \put(33, 99){\small{Ours}}
    \end{overpic}
    \vspace{-0pt}
    \caption{\textbf{SDF diffusion results.} We show ground-truth meshes and the input sparse point cloud (64 points) on the left. We compare our results with 3DS2VS. Since our model is probabilistic, we can output multiple different results given different random seeds. Our results are detailed and complete. However, the traditional method struggles to reconstruct correct objects.}
    \label{fig:sdf-results}
\end{figure}

We adapt the sampling method proposed in DDIM~\cite{song2020denoising} for the proposed functional diffusion. As shown in \cref{eq:noised}, the generating process is from timestep $t=1$ (\textcolor{source}{most noisy}) to $t=0$ (\textcolor{target}{least noisy}).
We start from an initial noise function $f_1 = \textcolor{source}{g}\in\mathcal{F}$.
Given the noised state $f_t$ at the timestep $t$, we obtain the ``less'' noised state $f_s$ where $0 \leq s < t \leq 1$,
\begin{equation}
    f_{s} = \alpha_s \underbracket{D_\theta[f_{t}, t]}_{\text{estimated $\textcolor{target}{f_0}$}} +  \sigma_s\underbracket{\left(\frac{f_t - \alpha_t D_\theta[f_{t}, t]}{\sigma_t}\right)}_{\text{estimated $\textcolor{source}{g}$}}.
\end{equation}
We can also write,
\begin{equation}\label{eq:sampling}
    f_{s}(\mathbf{x}) = \frac{\sigma_s}{\sigma_{t}}f_{t}(\mathbf{x}) + \left(\alpha_s - \sigma_s\frac{\alpha_{t}}{\sigma_{t}}\right) D_\theta\left(\{\mathbf{x}_i, f_t(\mathbf{x}_i)\}_{i\in\mathcal{C}}, t, \mathbf{x}\right)
\end{equation}

We sample a set $\mathcal{C}$ and evaluate $\{\mathbf{x}_i, f_t(\mathbf{x}_i)\}_{i\in\mathcal{C}}$ in every denoising step. 
The Eq.~\eqref{eq:sampling} shows how the one-step denoised function $f_s$ is obtained. We recursively apply the denoising process from $f_t$ to $f_s$.
In the end, we obtain the generated sample $\textcolor{target}{f_0}$.
More importantly, to obtain intermediate function values $f_s(\mathbf{x})$ for an arbitrary $\mathbf{x}$, we need to know $f_t(\mathbf{x})$, and thus all previous states for $\mathbf{x}$. However, when we are denoising the last step of the generation process, $\sigma_s = 0$, which means the generated function $f_0(\mathbf{x})$ is only dependent on the penultimate state of the function $\{\mathbf{x}_i, f_s(\mathbf{x}_i)\}_{i\in\mathcal{C}}$ (also see~\cref{fig:inference-chain}). With this observation, we can obtain the generated function values without knowing the intermediate states except the penultimate one. During inference, we only need to denoise the context set.
This is a key property of the proposed method which can accelerate the generation/inference. The sampling algorithm is summarized in~\cref{alg:sample}.

\begin{figure}[tb]
    \centering
    \begin{overpic}[trim={4cm 3cm 4cm 1cm},clip,width=\linewidth,grid=false]{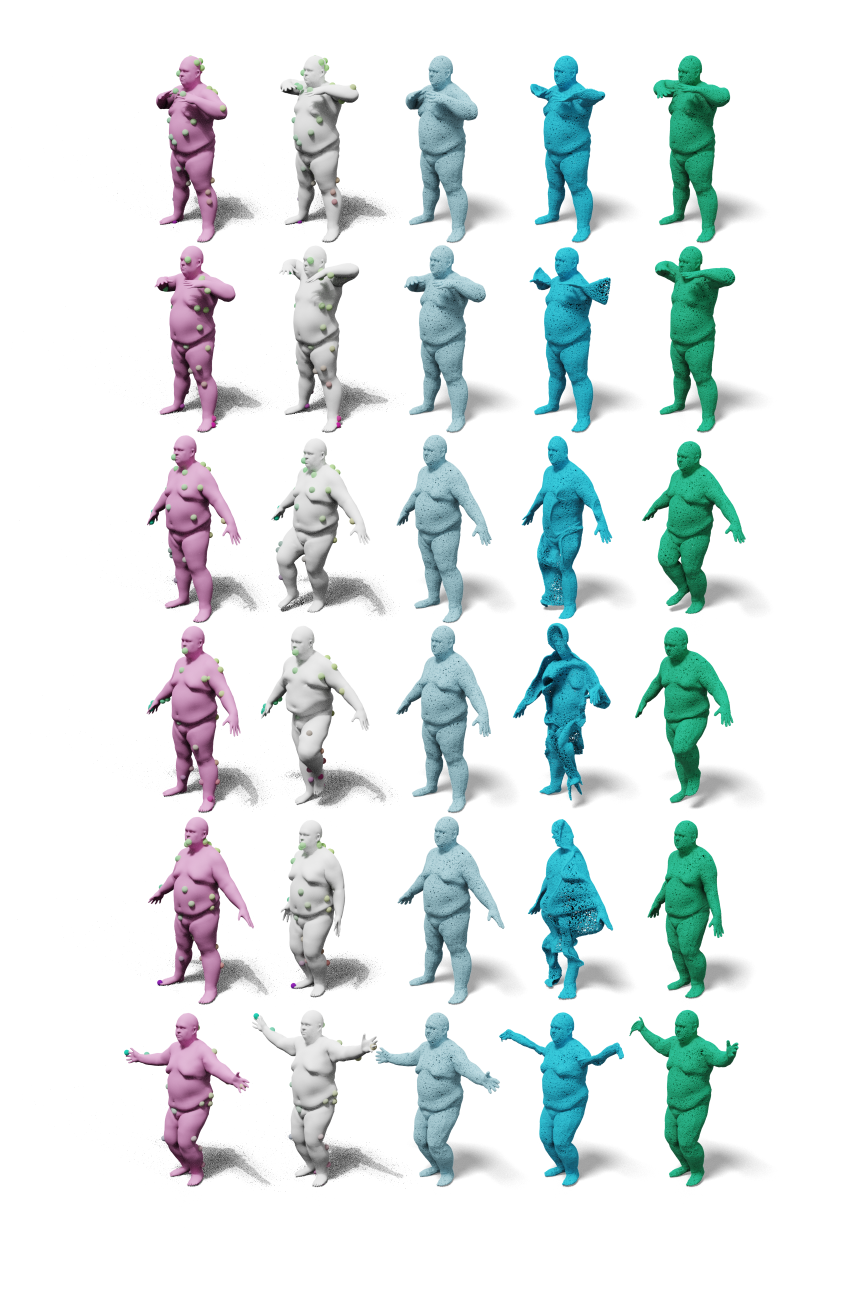}
        \dashline{0.7}(22,4)(22,98)
        \dashline{0.7}(33,4)(33,98)
        \dashline{0.7}(43,4)(43,98)
        \put(4, 98){\small{Source}}
        \put(14, 98){\small{Target}}
        \put(23, 98){\small{Queries}}
        \put(34, 98){\small{3DS2VS}}
        \put(46, 98){\small{Ours}}
    \end{overpic}
    \vspace{-0pt}
    \caption{\textbf{Deformation diffusion results.} In the left, we show both the source and the target frame and the sparse correspondence (small spheres on the body surface).}
    \label{fig:deform-results}
\end{figure}

\begin{figure*}[!htbp]
    \centering
    \begin{overpic}[trim={1cm 0cm 1cm -2cm},clip,width=\linewidth,grid=false]{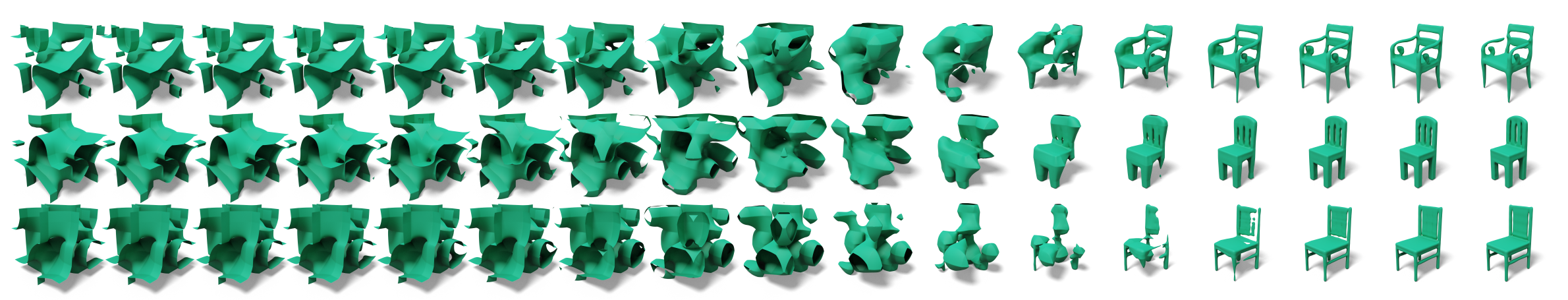}
        \put(0, 20){\small{Noise Function}}
        \put(87, 20){\small{Generated Function}}
        \put(15,20.5){\color{black}\vector(1,0){70}}
        \put(40, 21.5){\small{Generating Process}}
    \end{overpic}
    \vspace{-20pt}
    \caption{\textbf{Generating process of SDFs.} We show the generating process of 3 samples. In the far left, the initial noise functions are shown. In the far right, we show the generated samples. To make the visualization clear, we only show the zero-isosurface. However, the functions are actually densely defined everywhere in the space.}
    \label{fig:sdf-gen-process}
\end{figure*}

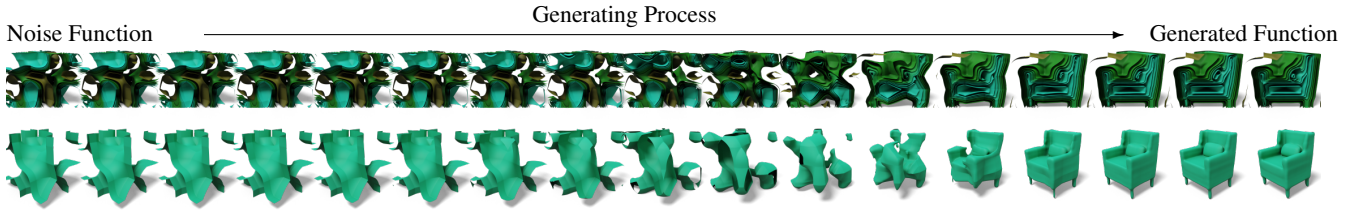
\begin{figure*}[!htbp]
    \centering
    \begin{overpic}[trim={1cm 0cm 1cm -2cm},clip,width=\linewidth,grid=false]{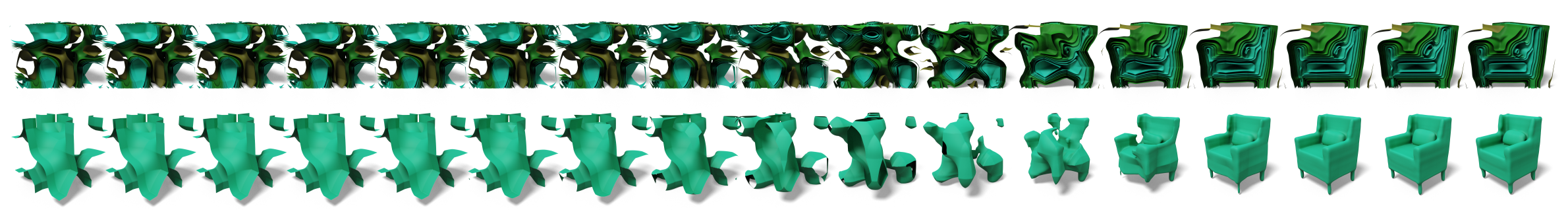}
        \put(0, 13){\small{Noise Function}}
        \put(87, 13){\small{Generated Function}}
        \put(15, 13.5){\color{black}\vector(1,0){70}}
        \put(40, 14.5){\small{Generating Process}}
    \end{overpic}
    \vspace{-20pt}
    \caption{\textbf{Generating process of SDFs.} In the top row, we show multiple isosurfaces of each intermediate step of the generating process. They are $[-0.5, -0.2, -0.1, -0.05, -0.01, 0, 0.05, 0.1, 0.2]$ from outer to inner. They are cut with a plane to show the inner structure. In the bottom row, we show the zero-isosurface for comparison.}
    \label{fig:sdf-gen-process-multi}
\end{figure*}

\section{Results: 3D Shapes}
In computer graphics, 3D models are often represented with a function $f: \mathbb{R}^3\rightarrow \mathbb{R}^1$
where the input $\mathbf{x}$ is a 3D coordinate and the output $y$ is the signed distance to the 3D boundary $\partial\Omega$, \ie, $y=\mathrm{dist}(\mathbf{x}, \partial\Omega)$ when $\mathbf{x}\in \Omega$ and $y=-\mathrm{dist}(\mathbf{x}, \partial\Omega)$ when $\mathbf{x}\in \mathcal{X}\setminus\Omega$.
The signed distance function (SDF) satisfies the partial differential equation (PDE), a.k.a., Eikonal equation,
\begin{equation}
    \begin{aligned}
        &\left|\nabla f(\mathbf{x})\right|=1, \\
        &\ f(\mathbf{x})=q(\mathbf{x}),\ \forall \mathbf{x}\in\partial\Omega ,
    \end{aligned}
\end{equation}
where $q(\mathbf{x})$ is the boundary condition. The task of predicting SDFs given a surface point cloud is equivalent to solving this PDE with a given boundary condition (the surface point cloud). This problem is solved in prior works. But most works focus on surface reconstruction only by predicting binary occupancies~\cite{mescheder2019occupancy, peng2020convolutional,yan2022shapeformer,zhang20223dilg} or truncated SDFs~\cite{park2019deepsdf}. Thus they are not really solving this equation and cannot be used in some SDF-based applications such as sphere tracing~\cite{hart1996sphere}. This is a challenging task according to prior works. We choose the task to show the capability of the proposed method.

\subsection{Experiment design}
We choose a sparse observation of the boundary condition (surface point cloud) which only contains $64$ points as the input of the model. We compare our method with OccNet~\cite{mescheder2019occupancy} and the recently proposed 3DShape2VecSet~\cite{10.1145/3592442}. As an example to show how the proposed method works, we first show how the noised state is obtained in~\cref{fig:noised_states}. The context is then fed into the denoising network (see~\cref{fig:sdf-pipeline}).

\subsection{Results comparison}
We show visual comparisons in~\cref{fig:sdf-results}. Apparently, our method shows a significant advantage over prior methods in this task. We not only output detailed and full meshes but also show the multimodality of the proposed method. However, prior works are unable to give correct reconstructions, thus also proving this task is challenging given the sparse observation.

We also show some quantitive comparison in~\cref{tab:sdf-quan}. Chamfer distances and F-scores are commonly used in surface reconstruction evaluation~\cite{mescheder2019occupancy, deng2020cvxnet, zhang20223dilg, 10.1145/3592442}. Furthermore, we design two new metrics. As discussed above, we are actually solving a partial differential equation. Thus, we can define the two metrics,
\begin{equation}
    \textsc{Eikonal}(f)=\frac{1}{|\mathcal
{E}_\mathcal{X}|}\sum_{i\in\mathcal{E}_\mathcal{X}}\left\|\left|\nabla f(\mathbf{x}_i)\right|-1\right\|^2,
\end{equation}
\begin{equation}
    \textsc{Boundary}(f)=\frac{1}{|\mathcal
{E}_\Omega|}\sum_{i\in\mathcal{E}_\Omega}\left\| f(\mathbf{x}_i)-q(\mathbf{x}_i)\right\|^2,
\end{equation}
where $\textsc{Eikonal}$ reflects that if the solutions satisfy the Eikonal equation and $\textsc{Boundary}$ shows if the solutions satisfy the boundary condition. $\mathcal
{E}_\mathcal{X}$ is a set sampled in the bounding volume which contains 100k points and $\mathcal
{E}_\Omega$ is a set sampled on the surface which also contains 100k points. Our method leads a large margin over existing methods in all metrics. This is also consistent with what is shown in the visual comparison.

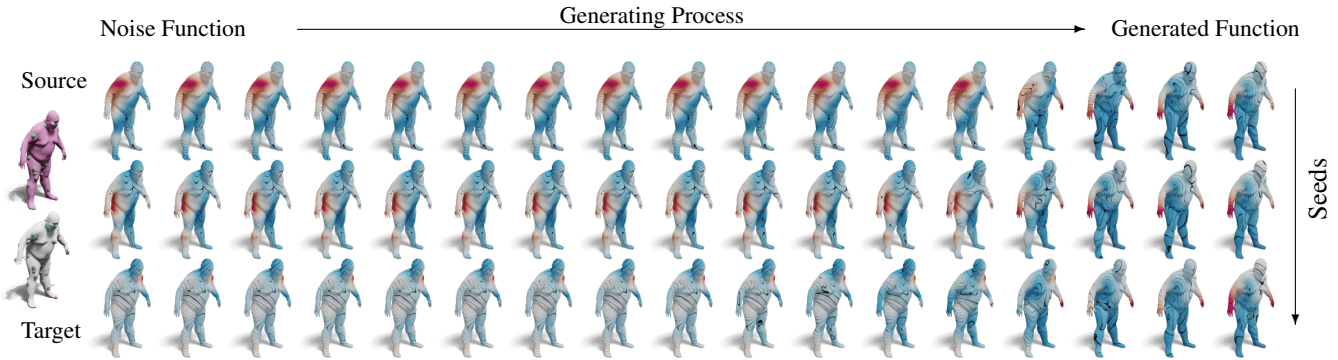
\begin{figure*}[!htbp]
    \centering
    \begin{overpic}[trim={3cm 2cm 3cm 0cm},clip,width=\linewidth,grid=false]{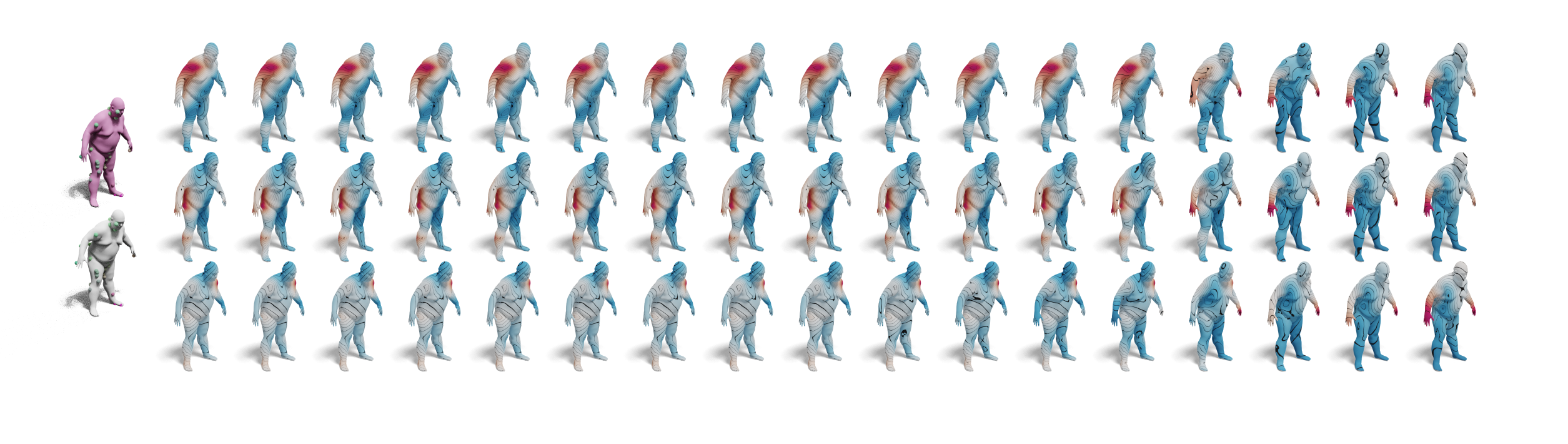}
        \put(8, 27){\small{Noise Function}}
        \put(85, 27){\small{Generated Function}}
        \put(23, 27.5){\color{black}\vector(1,0){60}}
        \put(43, 28){\small{Generating Process}}
        \put(2, 23){\small{Source}}
        \put(2, 4){\small{Target}}
        \put(99, 23){\color{black}\vector(0,-1){18}}
        \put(100,13){\rotatebox{90}{Seeds}}
    \end{overpic}
    \vspace{-20pt}
    \caption{\textbf{Deformation Fields.} In the far left, we show the source and target frame along with the sparse correspondence. We show 3 samples generated given the same condition. In each frame, we show the deformation field on the surface. However, to simplify the visualization, the colors only indicate the magnitudes of deformation while ignoring the directions. The three samples started from different functions. But in the end, the model outputs almost the same deformation fields.}
    \label{fig:deform-field}
\end{figure*}


\subsection{Generating process}
In~\cref{fig:sdf-gen-process} and~\cref{fig:sdf-gen-process-multi}, we show the intermediate noised function obtained during the generating process. Unlike similar methods proposed before which predict binary occupancies or truncated SDFs, we can generate raw SDFs directly which can be directly used in some SDF-based applications.

\section{Results: 3D Deformation}
The task is defined as follows: given meshes sampled in a dynamic shape sequence, and limited (32) sparse correspondence between two meshes (see~\cref{fig:deform-results}), we want to predict a deformation field. Specifically, the deformation field takes a point on the surface of the source frame as input and outputs a deformation vector which should map the point to the target frame. The network design is similar to the~\cref{fig:sdf-pipeline}. However, we only use 16384 points in the context set because the data is simpler than a complicated SDF.
We also adapt the method 3DS2VS here to do the deformation field prediction. From the visual results in~\cref{fig:deform-results}, we can see that our method can show vivid surface deformation, while 3DS2VS is unable to map source points to the target frame especially when the motion is large. We also show the quantitative comparisons in~\cref{tab:deforma-results}.

In~\cref{fig:deform-field}, we show what the generated deformation fields look like. Given the same condition, three sampling processes are visualized.

\begin{table}[tb]
\centering
\setlength{\tabcolsep}{3pt}
\begin{tabular}{ccccc}
\toprule
                        &  Chamfer $\downarrow$    &  F-Score $\uparrow$     & Boundary $\downarrow$ &  Eikonal $\downarrow$    \\ \midrule
OccNet & 0.166	& 0.531	& 0.019	& 0.032 \\
3DS2VS & 0.144 & 0.608 & 0.016 & 0.038 \\
Proposed& \textbf{0.101} & \textbf{0.707} & \textbf{0.012}& \textbf{0.024} \\\bottomrule
\end{tabular}
\caption{\textbf{SDF diffusion results.} The task is SDF prediction given sparse observations on the surface. We show two commonly used metrics, Chamfer distances and F-scores. Additionally, we show the two newly proposed metrics based on the definition of partial differential equations.}
\label{tab:sdf-quan}
\end{table}

\begin{table}[tb]
    \centering
    \begin{tabular}{cc}
    \toprule
        & MSE ($\times 10^4$) $\downarrow$ \\ \midrule
        3DS2VS & 13.32 \\
        Proposed & \textbf{6.91} \\ \bottomrule
    \end{tabular}
    \caption{\textbf{Quantitative results in deformation field generation.} The numbers are evaluated using minimum squared error between the predicted deformation and the ground-truth.}
    \label{tab:deforma-results}
\end{table}


\section{Conclusions}
We proposed a new class of generative diffusion models, called functional diffusion. In contrast to previous work, functional diffusion works on samples that are represented by functions. We derived the necessary foundations for functional diffusion and proposed a first implementation based on the transformer architecture.
\paragraph{Limitations.}
During our work, we identified two main limitations of our method. First, functional diffusion requires a fair amount of resources to train. 
However, other diffusion models also share the same issue.
We would expect that significantly more GPUs would be required to train on large datasets such as Objaverse-XL. Therefore, it may be interesting to explore cascaded functional diffusion in future work.
Second, our framework has an additional parameter, the sampling rate of the sampled function representation. During training, it is beneficial but also necessary to explore this hyperparameter.
\paragraph{Future works.}
In future work, we also would like to explore the application of functional diffusion to time-varying phenomena, such as deforming, growing, and 3D textured objects.
Furthermore, we would like to explore functional diffusion in the field of functional data analysis (FDA)~\cite{wang2016functional} which studies data varying over a continuum.

\clearpage
{
    \small
    \bibliographystyle{ieeenat_fullname}
    \bibliography{main}
}


\end{document}